\documentclass[runningheads]{llncs}
\usepackage{color}
\usepackage{graphicx}
\usepackage{amsmath}
\usepackage{amssymb}
\usepackage{booktabs}
\usepackage[misc]{ifsym}
\usepackage[
colorlinks,
linkcolor=blue,
anchorcolor=blue,
filecolor=blue,
urlcolor=blue,
citecolor=blue]{hyperref}
\usepackage[T1]{fontenc}
\begin{document}
\title{TeethDreamer: 3D Teeth Reconstruction from Five Intra-oral Photographs}
\titlerunning{TeethDreamer: 3D Teeth Reconstruction from Five Intra-oral Photographs}
%
\author{
Chenfan Xu\inst{1*} \and
Zhentao Liu\inst{1*}\and
Yuan Liu\inst{2} \and
Yulong Dou\inst{1} \and
Jiamin Wu\inst{3} \and
Jiepeng Wang\inst{1,2} \and
Minjiao Wang\inst{4} \and
Dinggang Shen\inst{1,5,6} \and
Zhiming Cui\inst{1}(\Letter)
}

\authorrunning{C. Xu et al.}
%
\institute{School of Biomedical Engineering \& State Key Laboratory of Advanced Medical Materials and Devices, ShanghaiTech University, Shanghai, China\\
\email{cuizhm@shanghaitech.edu.cn} \and
Department of Computer Science, The University of Hong Kong, Hong Kong, China \and
Applied Oral Sciences \& Community Dental Care, Faculty of Dentistry, The University of Hong Kong, Hong Kong, China \and
Shanghai Ninth People's Hospital, School of Medicine, Shanghai JiaoTong University, Shanghai, China \and
Shanghai United Imaging Intelligence Co. Ltd., Shanghai, China \and
Shanghai Clinical Research and Trial Center, Shanghai, China
}
%
\maketitle              

\begin{abstract}
Orthodontic treatment usually requires regular face-to-face examinations to monitor dental conditions of the patients.
When in-person diagnosis is not feasible, an alternative is to utilize five intra-oral photographs for remote dental monitoring.
However, it lacks of 3D information, and how to reconstruct 3D dental models from such sparse view photographs is a challenging problem.
In this study, we propose a 3D teeth reconstruction framework, named TeethDreamer, aiming to restore the shape and position of the upper and lower teeth.
Given five intra-oral photographs, our approach first leverages a large diffusion model's prior knowledge to generate novel multi-view images with known poses to address sparse inputs and then reconstructs high-quality 3D teeth models by neural surface reconstruction.
To ensure the 3D consistency across generated views, we integrate a 3D-aware feature attention mechanism in the reverse diffusion process.
Moreover, a geometry-aware normal loss is incorporated into the teeth reconstruction process to enhance geometry accuracy.
Extensive experiments demonstrate the superiority of our method over current state-of-the-arts, giving the potential to monitor orthodontic treatment remotely. Our code is available at \url{https://github.com/ShanghaiTech-IMPACT/TeethDreamer}

\end{abstract}
\section{Introduction}

\begin{figure}
\centering
    \includegraphics[width=0.98\textwidth]{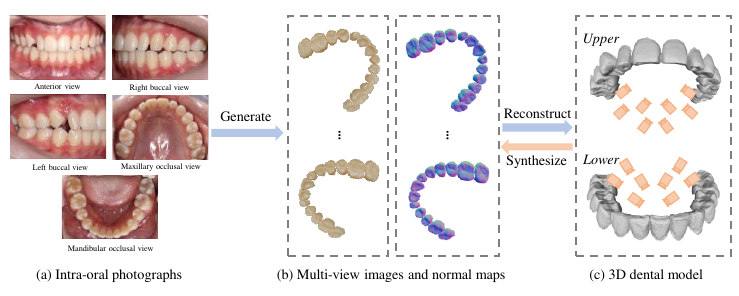}
    \caption{This flowchart illustrates our algorithm for reconstructing 3D dental models (c) from multiple intra-oral photographs (a). Initially, we synthesize multi-view images and normal maps (b) using the 3D dental model from our dataset. Subsequently, we train a diffusion network to generate these images and maps directly from the intra-oral photographs, culminating in the reconstruction of the target 3D dental models.
    } 
    \label{fig:intra-oral photos}
\end{figure}

Orthodontic treatment focuses on correcting teeth misalignment, such as malocclusion. This process typically extends over a long period, necessitating patients to regularly visit dentists for ongoing monitoring. While intra-oral scanning~\cite{intra-scanner} offers a way to acquire high-quality 3D dental models, it is often time-consuming and costly. In contrast, capturing several intra-oral photographs using smartphones presents a convenient alternative~\cite{intra-photo,intra-photo1}. Consequently, reconstructing 3D dental models from these 2D photographs for remote monitoring emerges as an attractive research direction.

The study referenced in~\cite{3Dtoothrecon} presents a method to reconstruct 3D dental structures from five intra-oral photographs using a parametric teeth model. While this approach yields decent results, it falls short in capturing personalized details of the teeth and the quality of the predefined teeth template also has a great effect on the reconstruction results. Besides, traditional Multi-View Stereo (MVS)~\cite{mvs} methods and recent implicit neural representation-based algorithms~\cite{neus,neus2,neuralangelo,neuralwrap} have achieved impressive reconstruction results in 3D vision society. However, these methods all require calibrated camera poses for accurate reconstruction, which is not feasible in our scenarios. Moreover, due to the sparsity and small overlap across intra-oral photographs, it is hard for current structure-from-motion (SfM)~\cite{sfm} methods to recover accurate camera poses. Recent successes in single image 3D reconstruction of natural objects~\cite{zero123,zero123++,wonder3d,syncdreamer,realfusion,one2345,one234++} have employed pretrained diffusion models~\cite{high-sd} to generate novel multi-view images from fixed viewpoints to reconstruct 3D objects. These generative methods generally condition on single view input to generate more images, which leads to inaccurate reconstruction of unseen regions. And particularly, they mostly generate color images without normal information, providing poor geometry information.

To address the limitations mentioned above, we propose a novel framework called TeethDreamer to reconstruct 3D teeth model only from five intra-oral photographs. Initially, we employ a pretrained diffusion model conditioned by segmented teeth images from intra-oral photos to generate multi-view color images and corresponding normal maps at specific viewpoints. These novel viewpoints help us to mitigate sparsity of input data in teeth reconstruction. To ensure consistency across different views, we further build 3D-aware feature from noisy color images and normal maps, and incorporate them into the diffusion model through an attention mechanism during the denoising process. Finally, we reconstruct the 3D teeth model through neural surface reconstruction with generated color images and normal maps. And a geometry-aware normal loss is introduced into the reconstruction process to improve the geometric accuracy. Extensive experiments have demonstrated our superiority over current state-of-the-arts, and give the potential to monitor orthodontic treatment remotely.

\section{Method}

\begin{figure}[t]
\centering
    \includegraphics[width=0.98\textwidth]{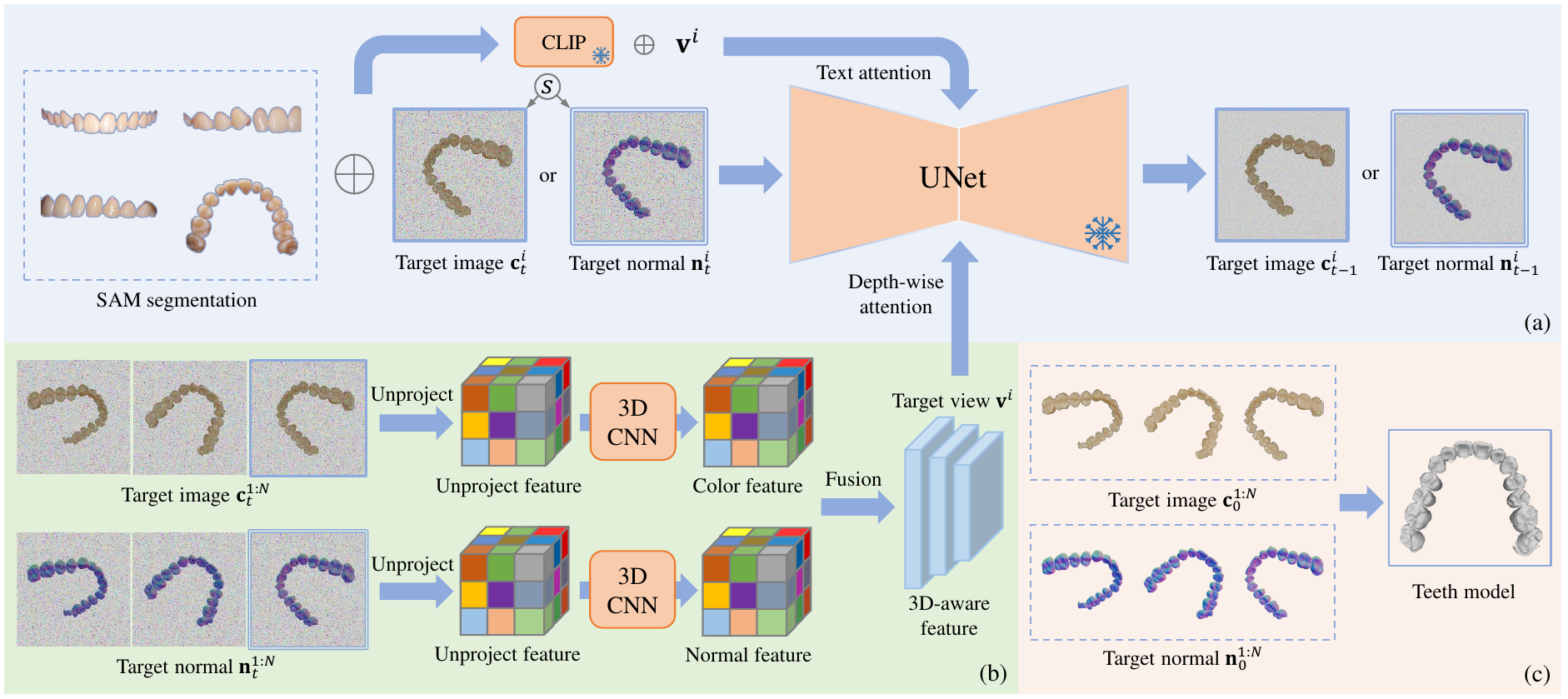}
    \caption{
    Overview of TeethDreamer. 
    (a) Generate color images and normal maps at different views from a pretrained diffusion model conditioned by segmented teeth images.
    Here, the diffusion model denoises the target view $\left\{\mathbf{c}^{i}_t,\mathbf{n}^{i}_t\right\}$ for one step.
    (b) 3D-aware feature extracted from all target views $\left\{\mathbf{c}^{1:N}_t, \mathbf{n}^{1:N}_t\right\}$ in latent domain to enforce consistency among generated views. 
    (c) Geometry-aware teeth reconstruction from generated color images and normal maps.
    } 
    \label{fig:pipeline}
\end{figure}

Given a set of intra-oral photographs, our goal is to reconstruct high-quality 3D models of upper and lower teeth. 
Our reconstruction framework has two stages. In the first stage, we train a diffusion model (Sec.~\ref{sec:diffusion}) to generate  multi-view consistent images and normal maps, along with a 3D-aware feature attention module to enforce multi-view consistency (Sec.~\ref{sec:consistent}). Then in the second stage, given the generated muti-view images and normal maps, we reconstruct 3D teeth via geometry-aware neural implicit surface optimization (Sec.~\ref{sec:reconstruction}). An overview of the proposed method is illustrated in Fig.~\ref{fig:pipeline}.


\subsection{Multiview Cross-domain Diffusion Model}
\label{sec:diffusion}
Given the intra-oral photographs, we first utilize the pretrained SAM model~\cite{SAM} to segment foreground teeth areas.
Note that, among five intra-oral photographs, one image (i.e., occlusal view) only contains the upper teeth or lower teeth.
As illustrated in Fig.~\ref{fig:pipeline}, we use four segmented images containing the upper teeth as model inputs, denoted as $\mathbf{x}^{1:4}$, and $\mathbf{x} \in \mathbb{R}^{3\times H\times W}$.

Due to the sparsity of input images, it is difficult to reconstruct high-quality 3D tooth models. Hence, we choose to augment the observing viewpoints with the help of generative diffusion models. Instead of RGB images, normal map is another important signal to recover 3D models. Therefore, we take segmented teeth image $\mathbf{x}^{1:4}$
as the input condition to a pretrained diffusion model $f$ from Zero123~\cite{zero123} to generate color images $\mathbf{c}^{1:N}$ and normal maps $\mathbf{n}^{1:N}$ at $N$ predefined viewpoints $\mathbf{v}^{1:N}$, denoted as:
%
\begin{equation}
\left(\mathbf{c}^{1:N},\mathbf{n}^{1:N}\right)=f\left(\mathbf{x}^{1:4},\mathbf{v}^{1:N}\right)
\label{eq:goal}
\end{equation}
Both $\mathbf{c}$ and $\mathbf{n}$ share the same dimension as $\mathbf{x}$. Note that the ground truth color images and normal maps are pre-synthesized from paired intra-oral scan models, as depicted in Fig.~\ref{fig:intra-oral photos}. In this way, we could leverage strong zero-shot generalization ability of diffusion prior. Besides, we could make use of the rich geometric information from normal maps to improve the teeth  reconstruction accuracy (will be described in Sec.~\ref{sec:reconstruction}).


We aim to learn the joint distribution of all these views $p_{\theta}\left(\mathbf{c}^{1:N}, \mathbf{n}^{1:N}|\mathbf{x}^{1:4}\right)$ which could be mathematically formulated into a multiview diffusion model. 
The reverse process could be simply extended from vanilla DDPM~\cite{ddpm} as follows.
\begin{equation}
p_{\theta}\left(\mathbf{c}^{1:N},\mathbf{n}^{1:N}|\mathbf{x}^{1:4}\right)=p\left(\mathbf{c}^{1:N}_T,\mathbf{n}^{1:N}_T|\mathbf{x}^{1:4}\right) \prod_{t=1}^{T} p_\theta\left(\mathbf{c}^{1:N}_{t-1},\mathbf{n}^{1:N}_{t-1} | \mathbf{c}^{1:N}_t,\mathbf{n}^{1:N}_t, \mathbf{x}^{1:4}\right)
\label{eq:chain}
\end{equation}
where $\left\{\mathbf{c}^{1:N}_t,\mathbf{n}^{1:N}_t\right\},t=0,1,...,T$ are latent variables.
As shown in Fig.~\ref{fig:pipeline}(a), we concatenate the input views $\mathbf{x}^{1:4}$ with noisy target view $\left\{ \mathbf{c}^{i}_t,\mathbf{n}^{i}_t \right\},i=1,...,N$ as input to the UNet.
Moreover, following Zero123, we also use
the attention layers of stable diffusion (text attention branch in Fig.~\ref{fig:pipeline}(a)) to process the concatenation of target view point $\mathbf{v}^{i}$ and the CLIP~\cite{clip} image features of the input views $\mathbf{x}^{1:4}$.

However, training the diffusion model to simultaneously generate color images and normal maps affects the performance of the pretrained model, due to discrepancy in the number of output channels. To address this, we employ a domain switcher $s\in\mathbb{R}^1$, which determines the output type, either color images or normal maps. In all, we train a multi-view cross-domain diffusion model and the formulation of Eq.~\ref{eq:goal} is modified as follows.
\begin{equation}
\mathbf{c}^{1:N},\mathbf{n}^{1:N}=f\left(\mathbf{x}^{1:4},\mathbf{v}^{1:N},s_c\right),f\left(\mathbf{x}^{1:4},\mathbf{v}^{1:N},s_n\right)
\label{eq:model}
\end{equation}

\subsection{3D-aware Feature Attention}
\label{sec:consistent}

Maintaining consistency across images and normal generated from various views is essential for high-quality geometry reconstruction. To achieve this, we introduce a method employing a 3D-aware feature extractor combined with a depth-wise attention mechanism. This strategy integrates the intermediate states $\left\{\mathbf{c}^{1:N}_t,\mathbf{n}^{1:N}_t\right\}$ during the denoising process for the current target view $\left\{\mathbf{c}^{i}_t,\mathbf{n}^{i}_t\right\}$. Initially, the generated 2D images $\mathbf{c}^{1:N}_t$ and normal maps $\mathbf{n}^{1:N}_t$ in the latent space are backprojected onto predefined 3D voxel grids with a size of $64^{3}$. A 3D CNN is then utilized to encode the color and normal feature volumes separately. Following this, a 3D U-Net merges these feature volumes, ensuring the outputs are consistent in both geometry and appearance. To extract features specific to the target viewpoint $\mathbf{v}^i$, we create a view frustum and perform interpolation within the resulting 3D feature volume. These view-specific features are then integrated into the denoising process through a depth-wise attention layer. This method effectively captures the spatial relationships between different views and consolidates essential information for target viewpoint, significantly improving consistency among the generated views.

\subsection{Geometry-aware Teeth Reconstruction}
\label{sec:reconstruction}

Due to the lack of camera parameters in intra-oral photos, we rely solely on generated color images and normal maps with predefined view points for teeth surface reconstruction based on Neus~\cite{neus,neus2}. 
Notably, we enhance this process by incorporating an additional geometry-aware normal loss. This allows us to extract high-quality 3D geometry from 2D normal maps with less noise.

Specifically, We first segment teeth masks $\mathbf{m}^{1:N}$ from generated color images $\mathbf{c}^{1:N}$ or or normal maps $\mathbf{n}^{1:N}$.
Then we randomly sample a batch of training pixels with associated rays $R_k=\{n_k, c_k, m_k, d_k\} \in \mathbf{R}$ from the train set $\{ \mathbf{n}^{1:N}, \mathbf{c}^{1:N}, \mathbf{m}^{1:N}\}$ for neural surface rendering.
Here, $n_k\in\mathbb{R}^3, c_k\in\mathbb{R}^3, m_k\in\{0,1\}, d_k\in\mathbb{R}^3$ represents the normal value, color value, mask value, and ray direction for $k_{th}$ ray, respectively.
The whole objective function is defined as follows.

\begin{equation}
\mathcal{L}=\mathcal{L}_{normal}+\mathcal{L}_{rgb}+\lambda_{1}\mathcal{L}_{mask}+\lambda_{2}(\mathcal{R}_{eik}+\mathcal{R}_{sparse})
\label{eq:loss}
\end{equation}
where $\mathcal{L}_{normal}$ denotes the normal loss term, which will be detailed later.
$\mathcal{L}_{rgb}$ measures the disparity between rendered color $\hat{c}_k$ and generated one $c_k$.
$\mathcal{L}_{mask}$ calculates the binary cross-entropy between rendered mask $\hat{m}_k$ and segmented one $\hat{m}_k$.
$\mathcal{R}_{eik}$ is the eikonal regularization aimed at ensuring the smoothness of reconstructed surface, while $\mathcal{R}_{sparse}$ is the sparsity regularization to reduce floaters.
$\lambda$ is the corresponding weight term for each loss term.

Now we delve into the normal loss that we have introduced.
Leveraging the differentiable nature of SDF representation in Neus, we can get the normal value of the inherently reconstructed surface by calculating second-order gradients of SDF. We adopt a geometry-aware normal loss to minimize the discrepancy between rendered normal $\hat{n}_k$ and reference one $n_k$:
\begin{equation}
\mathcal{L}_{normal}=\frac{1}{\sum w_k}\sum{w_k e_k},
~e_k=1-\mathrm{cos}(\hat{n}_k,n_k)
\label{eq:normal_loss}
\end{equation}
and $w_k$ is a geometry-aware weight defined as:
\begin{equation}
w_k=
\left\{
\begin{aligned}
&0,  &\mathrm{cos}(d_k, n_k)>\epsilon \\
&\mathrm{exp}\left(|\mathrm{cos}(d_k, n_k)|\right), &\mathrm{cos}(d_k,n_k)\leq \epsilon
\end{aligned}
\right
.
\label{eq:wk}
\end{equation}
here, $\epsilon$ is a negative threshold closing to zero.
The rationale behind such design lies in the fact that view direction $d_k$ is always opposite to the surface normal direction $n_k$, that is, the angle between them should always falls into the range of $[90^{\circ}, 180^{\circ}]$.
Any deviation from this condition would indicate inaccuracies of the generated normal $n_k$, thereby diminishing the effect in guiding the surface reconstruction process, i.e., $w_k=0$.

\section{Experiments}
\subsection{Experimental Settings}

\noindent\textbf{Dataset.}
We collected 3,200 cases to train our diffusion model, with each case including five intra-oral photos paired with an intra-oral scanning model. 
For the dataset, 3,000 cases were allocated for training, 100 for validation, and 100 for testing. 
We first segmented upper and lower teeth images from the intra-oral photos, and adjust them to $256\times256$ pixels via padding and downsampling.
We then synthesize target color images and normal maps from the intra-oral scans at eight specific viewpoints using Blenderproc~\cite{blenderproc}.
Each synthesized image has a resolution of $256\times256$.
For elevation angles, half the viewpoints for lower teeth are set at $45^{\circ}$ and the other half at $67.5^{\circ}$, with the opposite for upper teeth. Azimuth angles are divided into four groups: $30^{\circ}$, $60^{\circ}$, $120^{\circ}$, and $150^{\circ}$.

\noindent\textbf{Implementation Details.}
Our model training comprises two stages, including diffusion model fine-tunig and 3D teeth reconstruction.
In the first stage, the 3D CNN, 3D UNet, and attention layer are trainable, while the diffusion UNet and CLIP are frozen. 
This fine-tuning process takes 30k steps ($\sim$ 4 days) with a batch size of 64.
The learning rate starts with $1e^{-5}$ and linearly increases to $5e^{-4}$ by first 10k steps.
As for teeth reconstruction, we train the Neus model for 20k steps ($\sim$ 20 mins) with a ray batch size of $|\mathbf{R}|=4096$.
We primarily utilize normal and color information to supervise our geometry and appearance.
The mask information sharpens reconstruction edges and the remaining regularizations are auxiliary.
Thus the weights for different loss terms are set as follows:
$\lambda_1=0.1, \lambda_2=0.01$.
The learning rate initially increases from $1e^{-5}$ to $5e^{-4}$ by the first 500 steps in order to warm up training process, then undergoes exponential decay to $5e^{-5}$.
All experiments are conducted on a Single A100 GPU.

\noindent\textbf{Baselines and Metrics.}
We compare our approach with three baseline methods: Neus, Zero123, and SyncDreamer. 
Neus only uses teeth images segmented from intra-oral photos as inputs.
Both Zero123 and SyncDreamer leverage diffusion priors to generate novel color images from a single input.
The generated color images are then utilized to reconstruct the teeth models with Neus.
We chose maxillary and mandibular occlusal view as shown in Fig.~\ref{fig:intra-oral photos}(a) for their input, as these two views offer the most complete information of upper and lower teeth respectively.
To evaluate the accuracy of reconstructed teeth meshes, we calculate the Hausdorff Distance (HD), Chamfer Distance (CD), and IoU metrics against reference meshes.
Both HD and CD are quantified in millimeter (mm).
For assessing the quality of generated images, we measure PSNR, SSIM~\cite{ssim}, and LPIPS~\cite{lpips} metrics between ground truth and generated color images.

\begin{figure}[t]
\centering
    \includegraphics[width=0.98\textwidth]{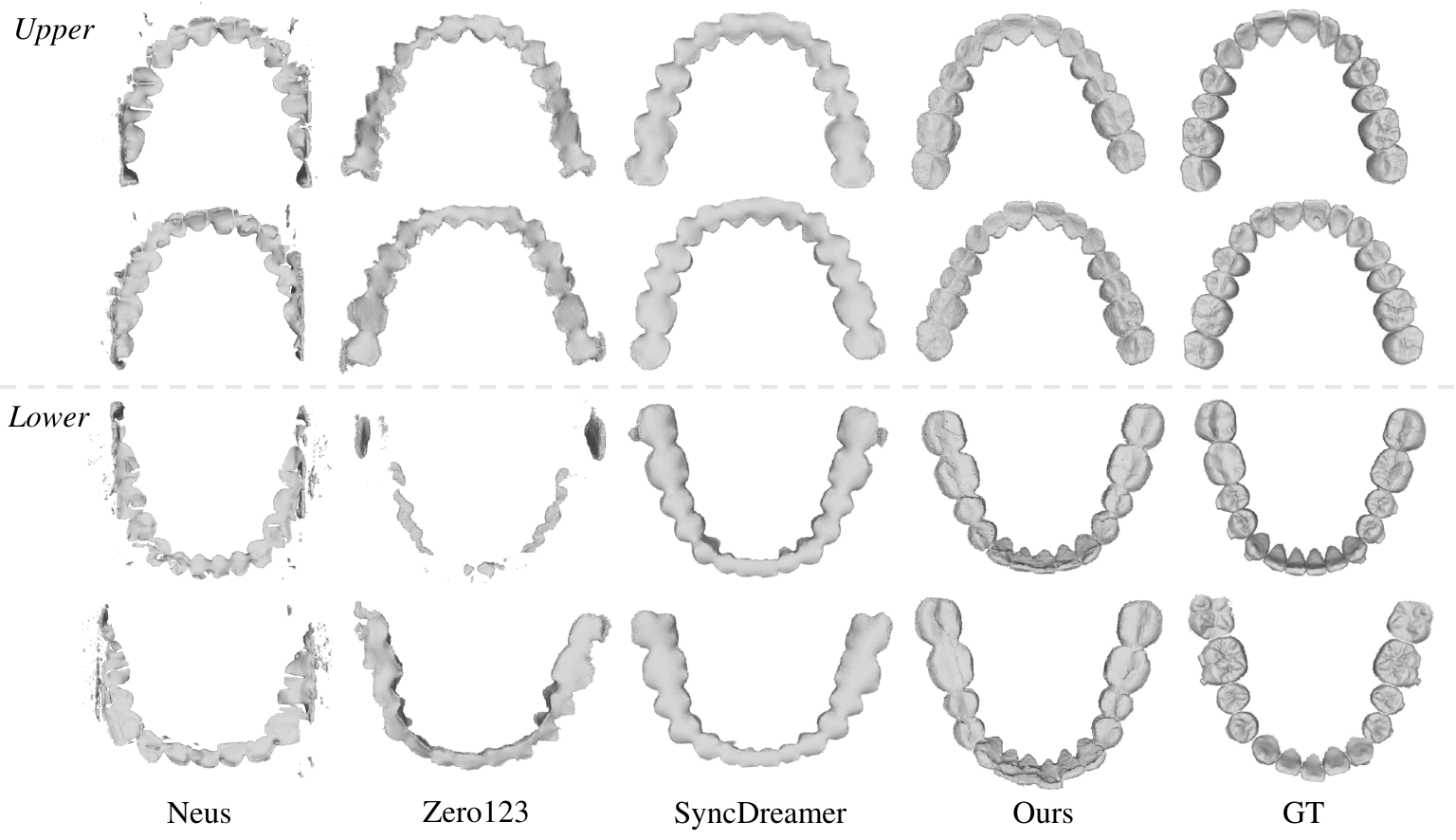}
    \caption{
    Qualitative comparisons of reconstructed 3D teeth with other baselines, demonstrating our results with complete shapes and geometric details. (GT: ground truth)
    } 
    \label{fig:comparison}
\end{figure}

\begin{figure}[t]
\centering
    \includegraphics[width=0.98\textwidth]{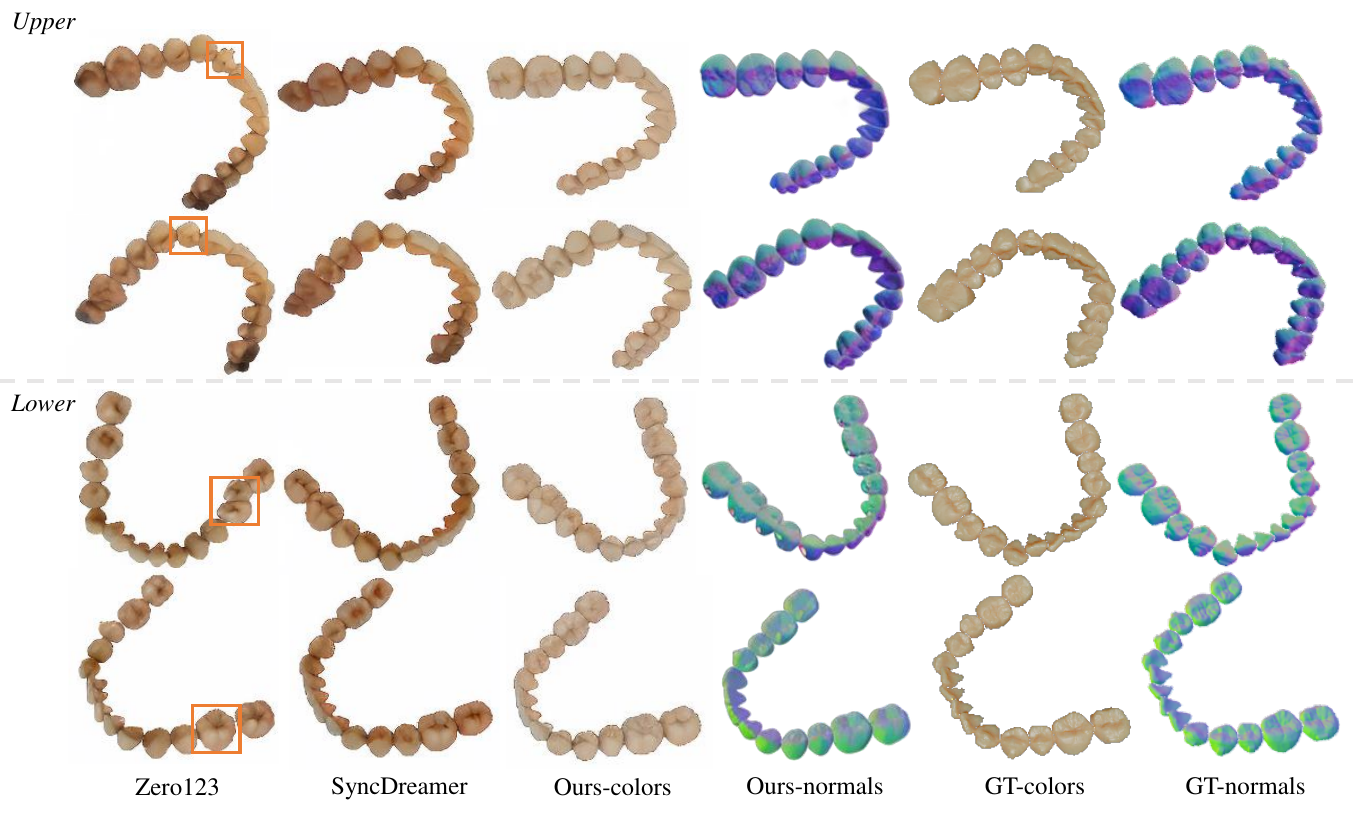}
    \caption{
    Qualitative comparisons of generated images with other baselines, where our generations are closely aligned with ground truth. (GT: ground truth)
    } 
    \label{fig:geneartion}
\end{figure}

\begin{table}[t]
  \begin{center}
    \caption{The quantitative comparison with other baselines and ablated solutions in color image generation and teeth reconstruction. }
    \label{tab:comparison}
    \centering
    \setlength{\tabcolsep}{5.5pt}
    \renewcommand\arraystretch{0.8}
    \begin{tabular}{lcccccc} 
    \toprule[1pt]
      \multicolumn{1}{c}{Method} & PSNR$\uparrow$ & SSIM$\uparrow$ & LPIPS$\downarrow$ & 
 HD(mm)$\downarrow$ & CD(mm)$\downarrow$ & IoU$\uparrow$\\
    \midrule[0.5pt]
      Neus~\cite{neus} & - & - & - & 3.8227 & 0.7770 & 0.0084\\
      Zero123~\cite{zero123} & 16.92 & 0.8123 & 0.0191 & 2.9281 & 0.2877 & 0.2446\\
      SyncDreamer~\cite{syncdreamer} & 17.19 & 0.8219 & 0.0170 & 2.5505 & 0.2045 & 0.3760\\
      \midrule[0.5pt]
      w/o 3D-aware & 16.63 & 0.8166 & 0.0171 & 4.2760 & 0.4371 & 0.1262\\
      w/o $\mathcal{L}_{normal}$ & - & - & - & 3.1760 & 0.2520 & 0.3521\\
      Ours & \textbf{18.85} & \textbf{0.8347} & \textbf{0.0114} & \textbf{2.1126} & \textbf{0.1670} & \textbf{0.4122}\\
      \bottomrule[1pt]
    \end{tabular}
  \end{center}
\end{table}

\subsection{Experimental Results}

Fig.~\ref{fig:comparison} illustrates the 3D teeth reconstruction results from TeethDreamer and other competing methods of two typical cases.
Neus suffers from sparse input with unknown poses, leading to poor reconstructions with floaters and distortions.
Zero123 fails to ensure consistency across generated views, resulting in incomplete meshes.
Although SyncDreamer restores the overall teeth shapes but it falls short in recovering fine-grained details.
In all, our results yield the best performance on reconstructed teeth with the help of 3D-aware feature attention and normal map guidance.
Fig.~\ref{fig:geneartion} demonstrates the comparison of generated images at two different viewpoints.
Zero123 fails to maintain consistency across different views, as highlighted in the orange box.
SyncDreamer struggles to accurately capture geometric details.
In contrast, our results closely align with the reference images in terms of both colors and normal maps, further demonstrating our effectiveness.
Table.~\ref{tab:comparison} depicts the quantitative analysis on 3D teeth reconstruction and 2D color image generation.
The performance trend is consistent with the qualitative results presented above.
Our method markedly outperforms other methods across all metrics.
Compared to SyncDreamer, we achieve an improvement of 2.22dB in PSNR and a reduction of 0.4379mm in HD, respectively.


\subsection{Ablation study}

In this section, we conduct ablation study to verify the effectiveness of two key components in our approach: 3D-aware feature attention and normal map guidance. 
Specifically, we study the effect of removing the 3D-aware feature attention (w/o 3D-aware) in diffusion model and the effect of omitting the geometry-aware normal loss (w/o $\mathcal{L}_{normal}$) in the teeth reconstruction.
The quantitative results are presented in Table.~\ref{tab:comparison}.
The results demonstrate that 3D-aware feature attention plays a vital role in producing high-quality 2D color images and in the reconstruction of 3D teeth model. Similarly, the geometry-aware normal loss is crucial for achieving high-quality 3D teeth reconstructions.

\section{Conclusion}
In this paper, we present a novel framework named TeethDreamer to reconstruct 3D teeth model from five intra-oral photographs.
We employ the diffusion prior to overcome input data sparsity, incorporated with 3D-aware feature attention mechanism to enhance view-consistency across generated views.
Moreover, a normal constraint is introduced into the teeth reconstruction process to increase geometric accuracy.
Extensive experiments have demonstrated our superiority over the current state-of-the-arts.
However, our reconstruction usually takes about 10~20 mins, which is time-consuming. 
The efficiency of neural surface reconstruction is one limitation of our framework which is also a wide concern in the filed of 3D reconstruction. 
Many researchers are working on speeding up it, such as e.g. instant-ngp~\cite{instant-ngp}, which we will consider to adapt to our task.
Although our method achieves the best performance currently, the fine-grained details are not always faithfully recovered. 
In future work, we plan to extract low-level features, such as edges or regions with high-frequency information, to obtain more precise details which may help improve the overall reconstruction quality.

%
%
%
%
\bibliographystyle{splncs04}
\bibliography{Paper-1038}

\end{document}